\documentclass[runningheads]{llncs}
\usepackage[T1]{fontenc}
\usepackage{graphicx}
\usepackage{csquotes}
\usepackage{subfig}
\usepackage{mathtools}
\usepackage{amsmath}
\usepackage{amssymb}
\usepackage{multirow}
\usepackage[misc]{ifsym}
\usepackage{algorithm}
\usepackage{algpseudocode}\usepackage{caption}
\begin{document}
\title{Few-shot Anomaly Detection in Text with Deviation Learning}
%
%
%
%
%
\author{Anindya Sundar Das(\Letter) \inst{1} \and
Aravind Ajay \inst{2} \and
Sriparna Saha \inst{2} \and 
Monowar Bhuyan \inst{1}}
\authorrunning{Das et al.}

\institute{Department of Computing Science,
Ume{\aa} University, Ume{\aa}, SE-90781, Sweden 
\email{\{aninsdas, monowar\}@cs.umu.se}
\and Department of Computer Science Engineering,
Indian Institute of Technology Patna, India\\
\email{\{aravindajay11, sriparna.saha\}@gmail.com}}
\maketitle              
\begin{abstract}
Most current methods for detecting anomalies in text concentrate on constructing models solely relying on unlabeled data. These models operate on the presumption that no labeled anomalous examples are available, which prevents them from utilizing prior knowledge of anomalies that are typically present in small numbers in many real-world applications. Furthermore, these models prioritize learning feature embeddings rather than optimizing anomaly scores directly, which could lead to \textit{suboptimal anomaly scoring} and \textit{inefficient use of data} during the learning process. In this paper, we introduce FATE, a deep few-shot learning-based framework that leverages limited anomaly examples and learns anomaly scores explicitly in an end-to-end method using deviation learning. In this approach, the anomaly scores of normal examples are adjusted to closely resemble reference scores obtained from a prior distribution. Conversely, anomaly samples are forced to have anomalous scores that considerably deviate from the reference score in the upper tail of the prior. Additionally, our model is optimized to learn the distinct behavior of anomalies by utilizing a multi-head self-attention layer and multiple instance learning approaches. 
Comprehensive experiments on several benchmark datasets demonstrate that our proposed approach attains a new level of state-of-the-art performance \footnote[1]{Our code is available at https://anonymous.4open.science/r/fate-1234/}.

\end{abstract}
%
%
%
\section{Introduction}

 Anomaly detection (AD) is defined as the process of identifying unusual data points or events that deviate notably from the majority and do not adhere to expected normal behavior. 
 Although the research on anomaly detection in the text domain is not particularly extensive, it has numerous relevant applications in Natural Language Processing (NLP) \cite{guthrie2007unsupervised,peng2018classifying,lee2021towards,crawford2015survey,ruff2019self}. 
 In general, anomaly detection faces two formidable obstacles. First, anomalies are dissimilar to each other and frequently exhibit distinct behavior. 
 Additionally, anomalies are rare data events, which means they make up only a tiny fraction of the normal dataset. Acquiring large-scale anomalous data with accurate labels is highly expensive and difficult; hence, it is exceedingly difficult to train supervised models for anomaly detection.
 Existing deep anomaly detection methods in text deal with these challenges by predominantly focusing on unsupervised learning, popularly known as one-class classification \cite{moya1993one,ruff2019self} or by employing self-supervised learning \cite{manolache2021date} in which detection models are trained solely on normal data. These models often identify noise and unimportant data points as anomalies \cite{aggarwal2017supervised}, as they do not have any previous understanding of what constitutes anomalous data. This results in higher rates of false positives and false negatives \cite{gornitz2013toward,ruff2019deep}.
 Nevertheless, these deep methods learn the intermediate representation of normal features independently from the anomaly detection techniques using Generative Adversarial Networks (GANs) \cite{deecke2019image,schlegl2017unsupervised} or autoencoders \cite{sakurada2014anomaly}. 
 As a result, the representations generated may not be adequate for anomaly detection.

 In order to counter these challenges, we propose a transformer based \textbf{F}ew-shot \textbf{A}nomaly detection in \textbf{TE}xt with deviation learning (\textbf{FATE}) framework in which we leverage a few labeled anomalous instances to train an anomaly-aware detection model, infusing a prior understanding of anomalousness into the learning objective. This setup is possible since only a small number of anomalous data points are required, which can be obtained either from a detection system that has already been established and validated by human experts or directly identified and reported by human users. 
 In this framework, we use a \textit{multi-head self-attention} layer \cite{lin2017structured} to jointly learn a vector representation of the input text and an anomaly score function. We optimize our model using a \textit{deviation loss} \cite{pang2019deep} based objective, where the anomaly scores of normal instances are pushed to closely match a reference score, which is calculated as the mean of a set of samples drawn from a known probability distribution (prior). Meanwhile, the scores for anomalous instances are forced to have statistically significant deviations from the normal reference score. FATE needs a tiny number of labeled anomalies for training, only  0.01\%-0.07\% of all anomalies in each dataset and just 0.008\%-0.2\% of total available training data. 

The \textbf{key contributions} of our work are listed below: i) We propose a new transformer-based framework for text anomaly detection based on few-shot learning that can directly learn anomaly scores end-to-end. Unlike existing approaches that focus on latent space representation learning, our model explicitly optimizes the objective of learning anomaly scores. To the best of our knowledge, this is the first effort to utilize a limited number of labeled anomalous instances to learn anomaly scores for text data. ii) Our proposed model employs various techniques, including multi-head self-attention, multiple-instance learning, Gaussian probability distribution and Z-score-based deviation loss to learn anomaly scores that are well-optimized and generalizable. We illustrate that our approach surpasses competing approaches by a considerable margin, especially in terms of sample efficiency and its capability to handle data contamination. iii) We conduct extensive evaluations to determine the performance benchmarks of FATE on three publicly accessible datasets. iv) Our study shows that FATE attains a new state-of-the-art in detecting anomalies in text data, as supported by extensive empirical results from three benchmark text datasets.

\section{Related Work}
Although there is a limited amount of literature on anomaly detection in text data, some important studies still exist. Our method is related to works from anomaly detection in text and few-shot anomaly detection.

Over the past few years, deep anomaly detection for images has attracted much attention with current works \cite{deecke2019image,wang2022cmg,zhang2021label,hendrycks2018deep} showcasing optimistic, cutting-edge outcomes.
A few methods for detecting text anomalies involve pre-trained word embeddings, representation learning, or deep neural networks. In one of the earlier studies \cite{manevitz2007one}, autoencoders are used for representation learning for document classification. Most of the recent works treat the task of anomaly detection in text data as a problem of detecting topical intrusions, which entails considering samples from one domain as inliers or normal instances. In contrast, a few samples from different disciplines are considered as outliers or anomalies. In one such study, CVDD \cite{ruff2019self} leverages pre-trained word embeddings and a self-attention mechanism to learn sentence representations by capturing various semantic contexts and context vectors representing different themes. The network then detects anomalies in sentences and phrases concerning the themes in an unlabeled text corpus. Recent works \cite{mai2022self,arora2021types,gangal2020likelihood,wu2022multi} use self-supervised learning to differentiate between types of transformations applied to normal data and learn features of normality. Token-wise perplexity is frequently used as the anomaly score for AD tasks. The DATE method \cite{manolache2021date} utilizes transformer self-supervision to learn a probability score indicating token-wise anomalousness, which is then averaged over the sequence length to obtain the sentence-level anomaly score for AD task.


The approaches mentioned above depend entirely on normal data to learn features of normality and they lack in utilizing the small number of labeled anomaly data points that are readily available. Therefore, the dearth of prior knowledge of anomalies often leads to ineffective discrimination by the models \cite{aggarwal2017supervised,gornitz2013toward,ruff2019deep}. Few-shot anomaly detection addresses the problem by utilizing a small number of labeled anomalous samples and therefore incorporating the prior knowledge of anomaly into the model. Injection of few-labeled anomalous samples into a belief propagation algorithm enhances the performance of anomalous node detection in graphs, as demonstrated in works such as \cite{mcglohon2009snare} and \cite{tamersoy2014guilt}. 
These techniques emphasize learning the feature space as the first step, which is then used to calculate the anomaly score. However, in contrast, Deviation Networks  are used in \cite{pang2019deep} and \cite{pang2021explainable} to efficiently leverage a small amount of labeled data for end-to-end learning of anomaly scores for both multivariate and image data, respectively. Multiple Instance Learning (MIL) \cite{sultani2018real,tian2021weakly,pang2021explainable} has also been investigated for anomaly detection in both image and videos in weakly supervised learning settings \cite{sultani2018real,tian2021weakly,pang2021explainable}. 
In this work, we introduce a transformer-based framework that leverages deviation learning and MIL methods in a novel and efficient way to learn scores reflective of text anomalousness.

\section{FATE: The Proposed Approach}

In this section, we explain the problem statement, recount the proposed FATE framework, and outline its various components.

\subsection{Problem Statement}
The objective of our FATE model is to employ a limited number of labeled anomalous samples along with a vast amount of normal data to train an anomaly-aware detection model that can explicitly learn anomaly scores. Given a training dataset $\mathcal{X} = \{ x_1, x_2, ..., x_{I}, x_{I+1}, x_{I+2}, ..., x_{I+O} \} $ consisting of $I$ normal data samples (inlier) $ \mathcal{X_I} = \{x_1, x_2, ..., x_I \}$,  a very small set of $O$ labeled anomalies (outlier) $\mathcal{X_O} = \{ x_{I+1}, x_{I+2}, ..., x_{I+O} \}$ with $I \gg O$ ($O$ is much smaller than $I$), which yields some insights into actual anomalies, we aim to learn an anomaly scorer $\psi_K: \mathcal{X} \rightarrow \mathbb{R}$ which can assign scores to data instances based on whether they are anomalous (outlier) or normal (inlier). The objective is to ensure $\psi_K(x_i)$ lies close to the mean anomaly scores of inlier data samples, defined by the sample mean from a prior distribution $\mathcal{N}(\mu ;\sigma)$, and $\psi_K(x_j)$ deviates significantly from the mean anomaly score and lies in the upper tail of the distribution, i.e., $\psi_K(x_j) > \psi_K(x_i)$, where $x_i \in \mathcal{X_I}$ and $x_j \in \mathcal{X_O}$.

\subsection{The Proposed Framework}
\begin{figure}[!h]
\centering
  \includegraphics[width=0.85\linewidth]{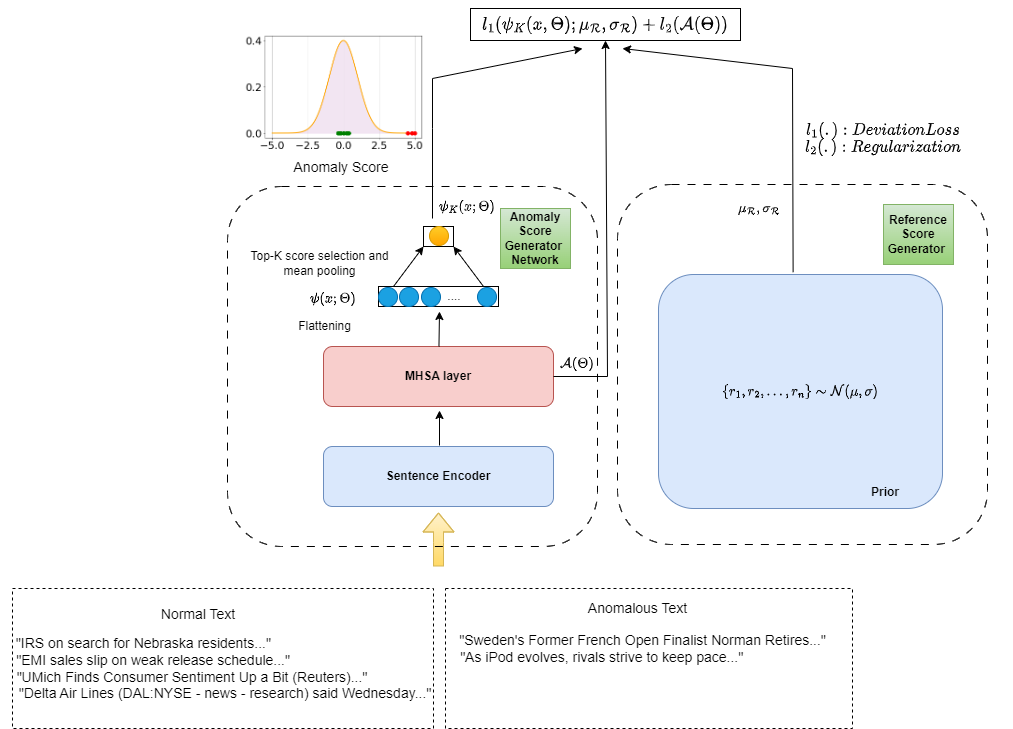}
  \caption{FATE: a proposed framework. The anomaly scoring network $\psi(x;\theta)$ is parameterized by $\Theta$, comprising the Sentence Encoder and Multi-Head Self-Attention (MHSA) layer. The attention matrix is denoted by $\mathcal{A}(\Theta)$. $\mu_{\mathcal{R}}$ and  $\sigma_{\mathcal{R}}$, are the mean and standard deviations of $n$-number of normal samples, respectively, which are defined by a prior probability distribution $\mathcal{N}(\mu;\sigma)$. $\psi_{K}$ denotes a top-K MIL-based anomaly scoring function. The loss $l_{1}(.)$ is deviation loss that ensures normal samples have anomaly scores close to $\mu_{\mathcal{R}}$, while scores for anomalous objects deviate significantly from $\mu_{\mathcal{R}}$. Loss $l_{2}(.)$ is the regularization term to enforce the attention heads to be nearly orthogonal.}
  \label{architec}
\end{figure}

The general architecture of our proposed FATE model is shown in Figure 
 \ref{architec}. It comprises two major components - an Anomaly Score Generator Network and a Reference Score Generator. The Anomaly Score Generator accepts an input text, a sequence of words of finite length, and processes it to produce a scalar output, i.e., anomaly score. The Reference Score Generator provides a reference for the anomaly score of normal samples. 

 \subsubsection{Anomaly Score Generator Network.}

The main components of the Anomaly Score Generator Network are \textit{Sentence Encoder}, \textit{Multi-Head Self-Attention (MHSA)} layer \cite{lin2017structured} and \textit{MIL-driven top-K scorer}.

\begin{itemize}

\item \textbf{Sentence Encoder.}  
Let $x_i=(x_{i1}, x_{i2},...,x_{iN})$ be an input text or token sequence comprising of $N$ words. The Sentence Encoder is a BERT-based \cite{devlin2018bert,reimers2019sentence} encoder that converts the input sequence $x_i$ into a sequence of contextualized representation $h(x_i)=(h_{i1},h_{i2},...,h_{iN}) \in \mathbb{R}^{d \times N}$, which is $d$-dimensional vector embedding representations of the $N$ tokens in $x_i$.


 \item \textbf{Multi-Head Self-Attention}. The MHSA layer accepts $h(x_i)$ as input and transforms it into vector embeddings $\psi(x_i;\Theta)$ of fixed-length $m$; each of the $m$ vectors represents a set of distinct anomaly scores, capturing multiple aspects and context of anomalousness in the text sequence, $x_i$. Formally, the MHSA layer computes an attention matrix $A = (a_1, a_2,...,a_m) \in (0,1)^{N \times m}$ from embedding representation $h(x_i) \in \mathbb{R}^{d \times N}$ as follows:

\begin{equation}
A = softmax(tanh(h(x_i)^{\intercal}\Theta_1)\Theta_2)
\end{equation}

\noindent where $\Theta_1 \in \mathbb{R}^{d \times r_a}$ and $\Theta_2 \in \mathbb{R}^{r_a \times m}$ are learnable weight matrices. The complexity of the MHSA module is determined by intermediate dimensionality $r_{a}$. Here the $softmax$ activation is applied on each column to ensure the resultant column vector $a_j$ of $A$ is normalized.  The $m$ vectors $a_1, a_2, ..., a_m$ are known as attention heads, and each head assigns relative importance to each token in the sequence.

The self-attention matrix $A$ is applied on embedding representation $h(x_i)$, which yields a fixed-length score matrix $S = (s_1, s_2, ..., s_m) \in \mathbb{R}^{d \times m}$ comprising of $m$ anomaly score vectors. Every column vector $s_k$ of $S$, that represents $d$ anomaly scores, is a convex combination of embedding vectors, $h_{i1},h_{i2},...,h_{iN}$.

\begin{equation}
S = h(x_i)A
\label{eq:score}
\end{equation}


To ensure independence between score vectors, eliminate redundant information and ensure that each score vector emphasizes on distinct aspects of semantic anomalousness, an orthogonality constraint called $MHSA\_loss$ is included in the model objective.
 
\begin{equation}
l_2 (x_i; \Theta) = || A^{\intercal}A - I ||^{2}_{F}
\label{eq:orth}
\end{equation}

\noindent where $||.||_{F}^2$ denotes the Frobenius norm \cite{golub2013matrix} of a matrix.

\item\textbf{MIL-driven top-K scorer.}
In Multiple-Instance-Learning (MIL) \cite{sultani2018real,tian2021weakly,pang2021explainable}, each data point is represented as a bag of multiple instances in different feature subspaces resulting in more generalized representations. The score matrix $S$ represents sets of orthogonal score vectors. The anomaly scoring vector $\psi(x_i;\Theta) \in \mathbb{R}^{\hat{N}}$, in which each input text is represented as a set of $\hat{N}$ distinct anomaly scores (multiple instances) is obtained after flattening $S$ (Fig. \ref{architec}). We then select $L(x_i)$, a set of K instances in $\psi(x_i;\Theta)$ with the highest anomaly scores. The overall MIL-driven top-K anomaly score of the input text is defined as:

\begin{equation}
\psi_K(x_i;\Theta) = \frac{1}{K}\sum_{x_{ij} \in L(x_{i})}
\psi(x_{ij};\Theta)
\end{equation}
\noindent where $|L(x_i)| = K$, $\psi(x_i;\Theta)= (\psi(x_{i1};\Theta), \psi(x_{i2};\Theta),..., \psi(x_{i\hat{N}};\Theta))$ and $\hat{N}=d*m$.

\end{itemize}
\subsubsection{Reference Score Generator.}

Once the anomaly scores have been obtained using $\psi_K(x_i;\Theta)$, the network output is augmented with a reference score $\mu_\mathcal{R} \in \mathbb{R}$ that provides prior knowledge to aid in the optimization process. This reference score is derived from the average anomaly scores of a set of $n$ randomly chosen normal instances, denoted as $\mathcal{R}$. 
To define the prior, we opt for a Gaussian distribution since \cite{kriegel2011interpreting} has provided extensive evidence that the Gaussian distribution is an excellent fit for anomaly scores across various datasets. We define a reference score based on Gaussian probability distribution:

\begin{equation}
    \mu_{\mathcal{R}}=\frac{1}{n}\sum^{n}_{j=1}r_i
    \label{eq:ref_score}
\end{equation}

\noindent where $r_i \sim \mathcal{N}(\mu, \sigma)$, denotes anomaly score of randomly selected $i$-th normal sample drawn from a Normal distribution and $\mathcal{R}=\{r_1, r_2, ..., r_n\}$ is prior-driven anomaly score set.

\subsubsection{Deviation Learning.} 
The deviation is determined as a Z-score with respect to the selected prior distribution:
\begin{equation}
    Z_{dev}(x_i; \Theta) = \frac{\psi_K(x_i;\Theta) - \mu_{\mathcal{R}}}{\sigma_{\mathcal{R}}}
\end{equation}

\noindent where $\sigma_{\mathcal{R}}$ is sample standard deviation of the set of reference scores $\mathcal{R}=\{r_1, r_2, ..., r_n\}$. The model learns the deviation between normality and anomalousness by optimizing a deviation-based contrastive loss \cite{hadsell2006dimensionality}, as given by Equation \ref{eqn:dev}.
\begin{equation}
    l_1(\psi_K(x_i;\Theta), \mu_{\mathcal{R}}, \sigma_{\mathcal{R}}) = (1-y_i)|Z_{dev}(x_i; \Theta)| + y_i\max(0, \alpha - Z_{dev}(x_i; \Theta))
    \label{eqn:dev}
\end{equation}

\noindent where $y_i=0$ if  $x_i \in \mathcal{X}_\mathcal{I}$ (normal sample set), and $y_i=1$ if $x_i \in \mathcal{X}_\mathcal{O}$ (anomalous sample set). The hyperparameter $\alpha$ in FATE is similar to the confidence interval parameter in Z-scores. By using this deviation loss, FATE aims  to minimize the difference between anomaly scores of normal examples and $\mu_{\mathcal{R}}$ while also ensuring that there is a minimum deviation of $\alpha$ between the anomaly scores of anomalous samples and $\mu_{\mathcal{R}}$. 

FATE optimizes both the deviation loss (Eqn. \ref{eqn:dev}) and the orthogonality constraint (Eqn. \ref{eq:orth}) objectives during the training process. The overall loss function is formulated as follows:

\begin{equation}
    l_{\mathcal{FATE}} (x_i; \Theta) =  l_1(\psi_K(x_i;\Theta), \mu_{\mathcal{R}}, \sigma_{\mathcal{R}}) + l_2 (x_i; \Theta)
\end{equation}

The procedure for training FATE is described in Algorithm \ref{alg:fate}.
FATE uses the optimized network $\psi_K$ during the testing phase to generate anomaly scores for each testing instance. The samples are ordered according to their anomaly scores, and those with the highest anomaly scores are identified as outliers. 
The Gaussian prior employed in generating the reference score makes our anomaly score inherently interpretable.

\begin{algorithm}
\caption{FATE: Training with few labeled anomalies}\label{alg:fate}
\hspace*{\algorithmicindent} \textbf{Input          :} $\mathcal{X}$: set of all training samples, $\mathcal{X_I}$: set of normal samples, $\mathcal{X_O}$: set of anomalous samples, $\mathcal{X}= \mathcal{X_I} \cup \mathcal{X_O}$ and  \O$= \mathcal{X_I} \cap \mathcal{X_O}$ \\
\hspace*{\algorithmicindent} \textbf{Output:} $\psi_K: \mathcal{X} \rightarrow \mathbb{R}$: Anomaly score generator
\begin{algorithmic}[1]

\State Initialize parameters in $\Theta$ randomly.
\For{$e=1$  $\rightarrow$ $number\_of\_epochs$} 
    \For{$b=1$  $\rightarrow$ $number\_of\_batches$}
        \State $\mathcal{X}_v$ 
        $\leftarrow$
        Select a random set of $v$ data instances consisting of an equal number of examples from both $\mathcal{X_I}$ and $\mathcal{X_O}$.
        \State Generate a set of $n$ anomaly scores $\mathcal{R}$ by sampling randomly from a normal distribution. $\mathcal{N}(\mu, \sigma)$
        \State Calculate the mean $\mu_{\mathcal{R}}$ and standard deviation $\sigma_{\mathcal{R}}$ from sample score set $\mathcal{R}$.
        \State Compute $deviation\_loss_{batch} = \frac{1}{b}\sum_{x \in \mathcal{X_b}} l_1(x,\mu_{\mathcal{R}}, \sigma_{\mathcal{R}}; \Theta)$.
        \State Compute $MHSA\_loss_{batch}= \frac{1}{b}\sum_{x \in \mathcal{X_b}} l_2(x;\Theta)$.
        \State Calculate total loss $loss_{batch} = deviation\_loss_{batch} + MHSA\_loss_{batch}$.
        \State Perform parameter optimization using Adam optimizer on $loss_{batch}$ $\Theta$.   
    \EndFor
\EndFor
\State return $\psi_{K}$.

\end{algorithmic}
\end{algorithm}

\section{Implementation Details}

We will first present here the datasets' specifics and subsequently provide information regarding the methodology used in the experimentation.

\subsection{Datasets} 

To assess how well FATE performs in relation to the standard baseline methods, we evaluate it on three benchmark datasets: 20Newsgroups \cite{lang1995newsweeder}, AG News \cite{zhang2015character}, and Reuters-21578 \cite{Lewis1997Reuters21578TC}. We preprocess them based on the methods suggested in \cite{ruff2019self}, \cite{manolache2021date}. 
This includes converting all text to lowercase, removing punctuation marks and numbers, and eliminating stopwords.
To prepare the training and testing splits for our few-shot setup, we use the following method for each dataset: Firstly, we create the inlier data (normal samples) by selecting examples from only one label of the train split of the corresponding dataset. Then we construct the outlier training set comprising a very small number of samples (5 to 40) by randomly sampling an equal number of instances from each outlier class (other labels in the train split).  When it comes to testing, we create the inlier data by selecting examples solely from one label of the test split of the corresponding dataset while using data with other labels in the test split as outlier test examples.
\begin{itemize}
    \item The \textbf{20Newsgroups} \cite{lang1995newsweeder} dataset consists of about 20,000 newsgroup documents grouped into 20 distinct topical subcategories. To conduct our experiments, we adhere to the approach outlined in \cite{manolache2021date} and \cite{ruff2019self} by only selecting articles from six primary classes: \textbf{computer, recreation, science, miscellaneous, politics,} and \textbf{religion}. 
    
    \item The \textbf{AG News} \cite{zhang2015character} dataset consists of articles gathered from diverse news sources assembled to classify topics. 
    
    \item \textbf{Reuters-21578} \cite{Lewis1997Reuters21578TC} is a well-known dataset in natural language processing that contains a collection of news articles published on Reuter's newswire in 1987. 
    Following \cite{ruff2019self}, we conduct experiments on $7$ selected categories: \textbf{earn, acq, crude, trade, money-fx, interest,} and \textbf{ship}.
\end{itemize}

\subsection{Model Training and Hyperparameters} 
We adhere to the process illustrated in Figure \ref{architec} to train the FATE network. We used PyTorch 1.12.1\footnote[2]{https://pytorch.org/} framework for all our experiments. For the sentence encoder, we fine-tune an SBERT model\footnote[3]{https://www.sbert.net/} and obtain the embeddings from the Encoder layer (before the pooling takes place). We employ a maximum sequence length of 128 and a batch size of 16. In the self-attention layer, we fix \textit{$r_a$} = 150 and \textit{m} = 5 for all experiments. Using an Adam optimizer with the learning rate set to $1\mathrm{e}{-6}$, we train our model for $5$ epochs on AG News. To account for the smaller dataset sizes of 20Newsgroups and Reuters-21578, we trained our model for $50$ epochs and $40$ epochs, respectively.  In some cases where the number of inlier sentences in the training data is lesser than $500$ (e.g., some subsets in Reuters-21578), we assign the number of epochs to $80$. K in the top-K scorer has been set to $10\%$, and the number of few-shot anomalies in each training has been set to 10. We have opted for the standard normal distribution $\mathcal{N}(0,1)$ as prior for reference score. The number of normal reference samples $n$ (Eqn. \ref{eq:ref_score}) has been set to 5000. We have selected a confidence interval $\alpha$ = $5$ (Eqn. \ref{eqn:dev}).

\subsection{Baselines and Evaluation Metrics}
We compare our proposed method FATE with existing state-of-the-art text AD models such as OCSVM\footnote[4]{OCSVM and CVDD implementations are taken from official CVDD work: https://github.com/lukasruff/CVDD-PyTorch} \cite{scholkopf2001estimating}, CVDD\footnotemark[4]\cite{ruff2019self} and DATE\footnote[5]{The experimentation is conducted using the official DATE published code and setup: https://github.com/bit-ml/date} \cite{manolache2021date}. 

In our experiments, we utilize the commonly used performance metrics for detection, which are the Area Under Receiver Operating Characteristic Curve (AUROC) and Area Under Precision-Recall Curve (AUPRC). AUROC metric describes the true positive versus false positive ROC curve, where AUPRC characterizes compact representation of precision-recall curve. 
Higher values for both AUROC and AUPRC correspond to better performance.

\section{Results and Analysis}
In this section, we will compare our model's performance against other state-of-the-art methods and then comprehensively analyse our proposed methodology from various perspectives to gain a profound understanding of its capabilities.
\paragraph{\textbf{Main Results.}}
{\small\tabcolsep5pt 
\begin{table}[]
    \caption{AUROC (in \%) and AUPRC (in \%) results of all the baselines and proposed models on 20Newsgroups, AG News, and Reuters-21578 are reported below. The reported values are the mean results obtained from multiple runs. }
    \centering
    \resizebox{\linewidth}{!}{
    \begin{tabular}{|c|c|c|c|c|c|c|c|c|c|}
    \hline
    \multirow{2}{*}{\textbf{Datasets}} & \multirow{2}{*}{\textbf{Inlier       }} & \multicolumn{2}{|c|}{\textbf{OCSVM       }} & \multicolumn{2}{|c|}{\textbf{CVDD      }} & \multicolumn{2}{|c|}{\textbf{DATE       }} & \multicolumn{2}{|c|}{\textbf{FATE      }} \\ \cline{3-10}
    & & AUROC & AUPRC & AUROC & AUPRC & AUROC & AUPRC & AUROC & AUPRC \\
    \hline
    \multirow{6}{*}{\textbf{20 News}} & comp & 78.0& 88.5 & 74.0& 85.3& 92.1& 96.8&\textbf{92.7} & \textbf{97.1}\\ \cline{2-10} 
    & rec & 70.0& 87.3 & 60.6 & 81.8 & 83.4& 93.5& \textbf{89.7} & \textbf{96.6}\\ \cline{2-10} 
    & sci &64.2 & 86.0 & 58.2& 82.4& 69.7& \textbf{89.9}& \textbf{73.7}& 89.5\\ \cline{2-10} 
    & misc &62.1& 96.2 & 75.7& 97.3& 86.0& 97.3& \textbf{89.2} &\textbf{99.3}\\ \cline{2-10} 
    & pol &76.1& 93.5 & 71.5& 91.8& 81.9& 95.8& \textbf{89.5} & \textbf{97.0}\\ \cline{2-10} 
    & rel &78.9& 95.0 &78.1& 94.1& 86.1& 96.8& \textbf{87.5} & \textbf{97.4}\\ \hline
    \multirow{4}{*}{\textbf{AG News}} & business &79.9& 93.0 & 84.0 & 93.2& 90.0 & 94.7 & \textbf{90.8} & \textbf{96.1}\\ \cline{2-10} 
    & sci & 80.7& 91.2 & 79.0 & 88.0& 84.0 & 92.9& \textbf{89.5} & \textbf{96.7} \\ \cline{2-10} 
    & sports & 92.4 & 97.3 & 89.9 & 95.4& 95.9 & 97.8& \textbf{99.2} & \textbf{99.7}\\ \cline{2-10} 
    & world & 83.2 &91.9 & 79.6 & 88.5&  90.1 & 95.0& \textbf{93.3} &\textbf{96.5} \\ \hline
     \multirow{7}{*}{\textbf{Reuters-21578}} & earn & 91.1 & 87.2 & 93.9 & 88.7& 97.4 &  97.5 & \textbf{97.6} & \textbf{98.0} \\ \cline{2-10} 
    & acq  & 93.1 & 95.7 & 92.7 & 95.2& 93.5 & 95.0 & \textbf{98.5} & \textbf{99.0}\\ \cline{2-10} 
    & crude  & 92.4 & 99.1 & \textbf{97.3} & 99.3& 78 & 97.7& 95.8 & \textbf{99.6}\\ \cline{2-10} 
    & trade  & 99.0 & \textbf{99.9} &\textbf{99.3} & \textbf{99.9}& 95.0 & 99.7& 97.8 & \textbf{99.9} \\ \cline{2-10} 
    & money-fx  & 88.6 & 99.3 & 82.5 & 99.0 & 88.3 & 98.7& \textbf{94.2} & \textbf{99.7}\\ \cline{2-10}
    & interest  & 97.1 & \textbf{99.9}& 95.9 & 99.8& 93.0& 98.8& \textbf{97.4} & \textbf{99.9} \\ \cline{2-10} 
    & ship  & 91.2& 99.7& 96.1 & 99.8& 77.7& 98.9& \textbf{96.5} & \textbf{99.9}\\ \hline

    \end{tabular}
    }
    
    \label{tab:result}
\end{table}
}

The outcomes of our proposed technique and those of previous state-of-the-art approaches on all datasets are shown in Table \ref{tab:result}. The AUROC values show that our method outperforms the top-performing DATE model by 3.85 and 3.2 percentage points on 20Newsgroups and AG News, respectively. In addition, our proposed technique surpasses CVDD by a significant margin of 17.4 points and 10.1 points on 20Newsgroups and AG News, respectively. Furthermore, for the Reuters-21578 dataset splits, our approach outperforms the best-performing CVDD method by 2.9 points and DATE model by 7.8 points. This clearly illustrates the efficacy of our proposed method.

\paragraph{\textbf{Ablation Study.}}
To validate the contributions of different components in the proposed approach, here we introduce four variants for ablation study: i) \textbf{\textit{FATE without Top-K:}} We remove the Top-K MIL layer and obtain anomaly score by averaging the outputs of the MHSA layer. ii) \textbf{\textit{FATE without MHSA:}} In this variant, the MHSA layer is removed, and the output from the sentence encoder is directly fed to the Top-K layer for anomaly score computation iii) \textbf{\textit{FATE with BCELoss:}} We derive this variant by replacing deviation loss with BCE loss
iv) \textbf{\textit{FATE with FocalLoss:}} In this implementation, focal loss \cite{lin2017focal} substitutes deviation loss.

Table \ref{tab:ablation} presents the results of the ablation study. We observe that removing the top-K layer significantly affects the model's performance on some datasets. Removal of the MHSA layer impacts the model's performance to some extent; moreover, it adversely impacts the overall stability of the model. We also found that the FATE variant with deviation loss performs much better on average than its BCE loss and Focal loss counterparts.
{\small\tabcolsep5pt 
\begin{table}[]
\caption{ AUROC (in \%) performance of the Ablation Study}
    \centering
    \resizebox{\linewidth}{!}{
    \begin{tabular}{|c|c|c|c|c|c|c|}
    \hline
    \textbf{Datasets} & \textbf{Inlier} & 
    \begin{tabular}[c]{@{}c@{}}FATE\\ without Top-K\end{tabular} &
    \begin{tabular}[c]{@{}c@{}}FATE\\ without MHSA\end{tabular} &
    \begin{tabular}[c]{@{}c@{}}FATE\\ with BCELoss\end{tabular} &
     \begin{tabular}[c]{@{}c@{}} FATE\\ with FocalLoss\end{tabular}
    & FATE\\ 
    \hline
    \multirow{6}{*}{\textbf{20 News}} & comp & 81.1	&90.0	&\textbf{92.7}	&91.6	&\textbf{92.7} \\ \cline{2-7} 
    & rec &84.6	& 86.3	& 83.2	& 82.9	& \textbf{89.7}
 \\ \cline{2-7} 
    & sci &67.2	&70.2	&71.4	&72.5
& \textbf{73.7} \\ \cline{2-7} 
    & misc &84.2	&87.3	&89	&88.4& \textbf{89.2} \\ \cline{2-7} 
    & pol &84.6	&84	&88.4	&84.9	& \textbf{89.5} \\ \cline{2-7} 
    & rel &85.1	&86.1	&87.4	&84.5	& \textbf{87.5} \\ \hline
    \multirow{4}{*}{\textbf{AG News}} & business &84.7	&88.7	&87.6	&84.7& \textbf{90.8} \\ \cline{2-7} 
    & sci & 78	&88.6	&84.7	&87.1 &\textbf{89.5} \\ \cline{2-7} 
    & sports & 94.5	& 97.5	&99	&98.8 & \textbf{99.2} \\ \cline{2-7} 
    & world & 71	&92.9	&\textbf{93.5}	&92.1& 93.3 \\ \hline
     \multirow{7}{*}{\textbf{Reuters-21578}} & earn & 89.3	&96.8	&98.1	&96.9 & \textbf{97.6}\\ \cline{2-7} 
    & acq  & 97.7	&97.8	&98.3	&98.1 & \textbf{98.5} \\ \cline{2-7} 
    & crude  & 85.5	&94.1	&92.4	&90.9	&\textbf{95.8}
 \\ \cline{2-7} 
    & trade  & \textbf{98.3}	&96.4	&98.2	&97.8 & 97.8 \\ \cline{2-7} 
    & money-fx  & \textbf{97.2}	&92.9	&94.1	&92.3& 94.2 \\ \cline{2-7}
    & interest  & 96.2	&96.1 &	\textbf{99.5}&	98.8& 97.4 \\ \cline{2-7} 
    & ship  & 94.2	&95.6	&\textbf{98.2}	&91.1& 96.5 \\ \hline

    \end{tabular}
    }
    
    \label{tab:ablation}
\end{table}
}

\begin{figure*}[!h]
  \centering
  \begin{minipage}[b]{0.45\textwidth}
    \includegraphics[width=\textwidth]{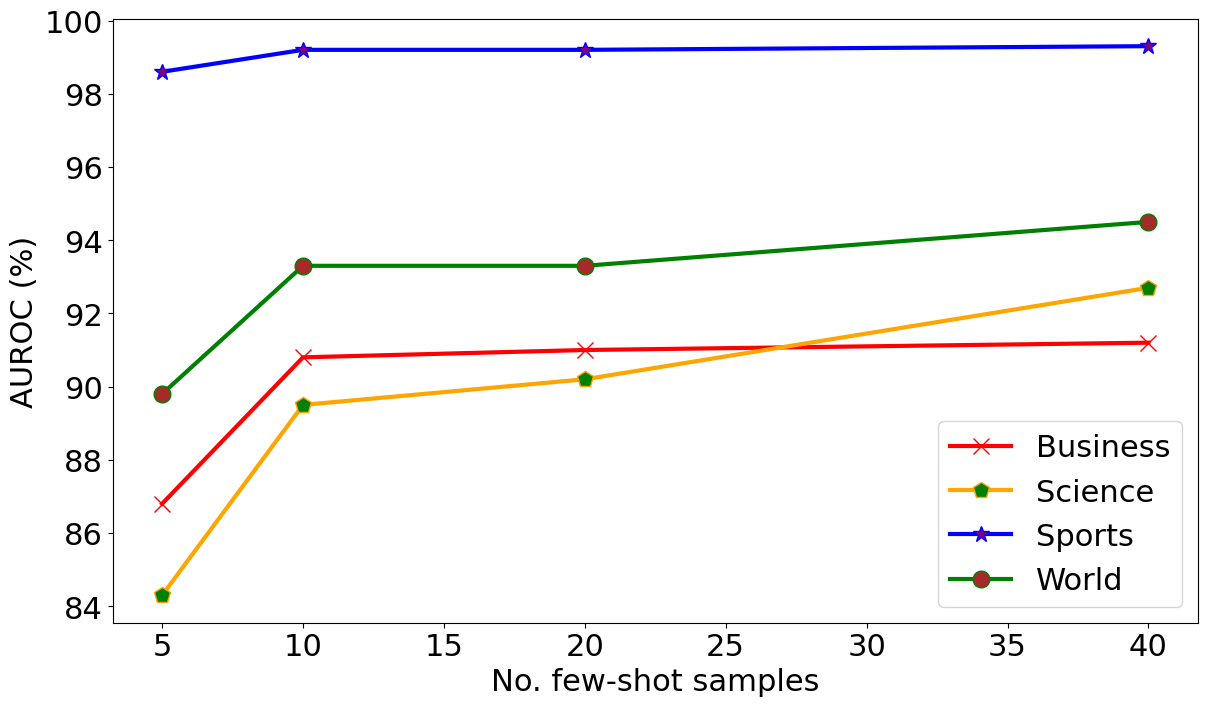}
    \caption{Sample efficiency: AUROC vs. no of labeled outlier samples for few-shot learning on AG News}
    \label{fig:sample}
  \end{minipage}
  \hfill
  \begin{minipage}[b]{0.45\textwidth}
    \includegraphics[width=\textwidth]{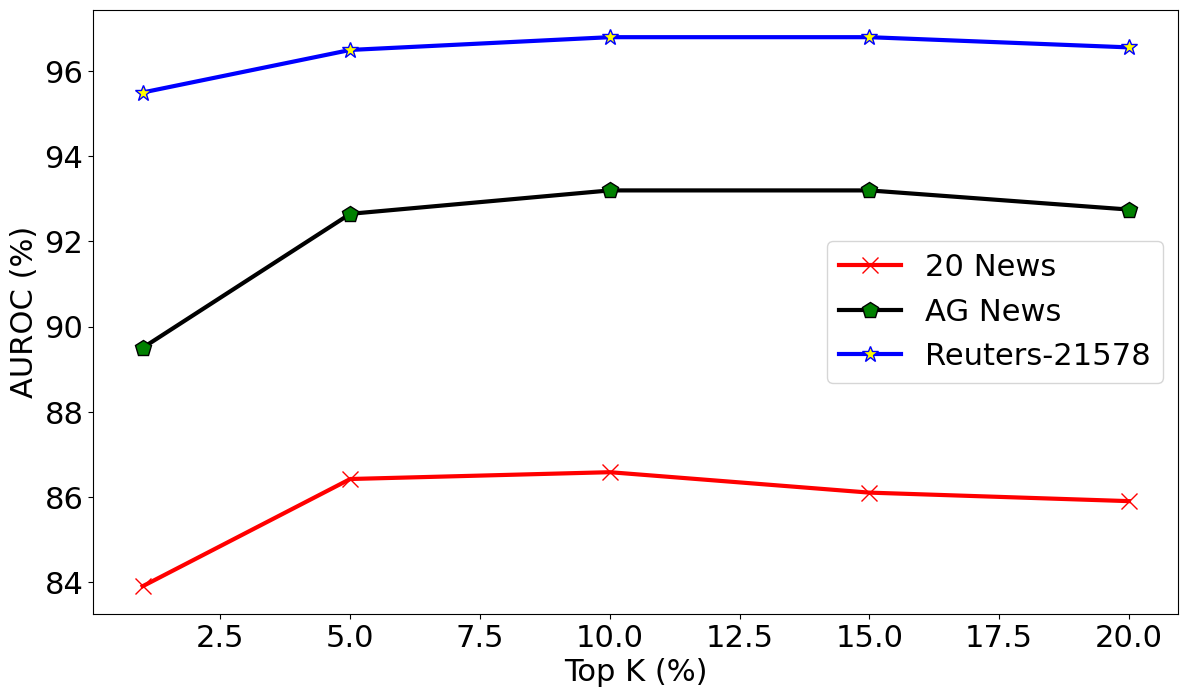}
    \caption{Sensitivity: AUROC performance with respect to Top-K (\%)}
    \label{fig:topk}
  \end{minipage}
  \includegraphics[width=0.65\linewidth]{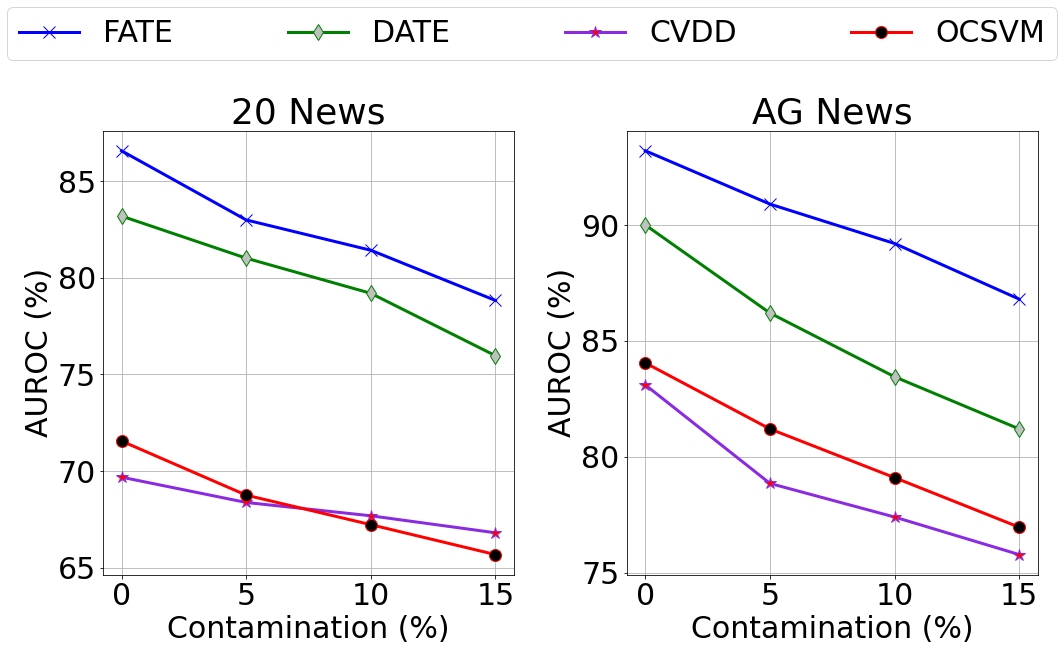}
    \caption{AUROC vs. Different Anomaly Contamination (\%) in inlier training data. The experiments are conducted on 20 News and AG News datasets.}
    \label{fig:cont}
\end{figure*}

 We also investigate the top-K MIL module by varying the value of K. As shown in Figure \ref{fig:topk}, the FATE model generally achieves improved performance as K increases from 1\% to 10\%. However, it is observed that the performance starts to decline on some datasets beyond this value of K. We find K=10\% works well on most of the datasets.

\paragraph{\textbf{Sample Efficiency.}}

To determine the optimal number of labeled anomalies needed to train FATE, we vary the number of labeled outliers from 5 to 40 in the training dataset. In contrast, the test data is remains unaltered. As shown in Figure \ref{fig:sample}, the FATE model achieves remarkable performance with only 10-20 labeled outlier samples, which is a very small fraction (0.03-0.06\% on AG News) of the available inlier training samples. This suggests that the FATE model is capable of efficiently utilizing labeled samples. Nonetheless, we also observe that the model's performance is affected when there are no representative samples in the outlier set from one of the outlier classes, highlighting the limitation of FATE in managing previously unseen classes.

\paragraph{\textbf{Robustness.}}
We further investigate the robustness of FATE in scenarios where the normal samples are contaminated with outliers. In this setup, anomalies are injected into normal training data as contamination. We compare our model's performance with other baselines. As shown in Figure \ref{fig:cont}, all methods experience a decline in performance as the contamination rate increases from $0\%$ to $15\%$. However, our proposed method is impressively resilient, as it outperforms the best-performing DATE model by $5.6\%$ on AG News and by $2.8\%$ on 20 News even when the contamination rate is as high as $15\%$.

\section{Concluding Remarks}
In this paper, we proposed FATE, a new few-shot transformer-based framework for detecting anomalies in text that uses a sentence encoder and multi-head self-attention to learn anomaly scores end-to-end. This is done by approximating the anomaly scores of normal samples from a prior distribution and then using a deviation loss that relies on Z-scores to push the anomaly scores of outliers further away from the prior mean. Evaluation on three benchmark datasets demonstrated that FATE significantly outperforms state-of-the-art methods, even in the presence of significant data contamination, and has a high level of efficacy in using labeled anomalous samples, delivering exceptional results even with minimal examples.  In our future work, we aim to investigate the effectiveness of FATE in detecting anomalies in text data when dealing with unseen classes and also aim to investigate the interpretability of FATE to help explain its decision-making process. This will help make it more transparent and understandable for end-users.


%
%
%

\subsection*{Acknowledgements}
This work was partially supported by the Wallenberg AI, Autonomous Systems and Software Program (WASP) funded by Knut and Alice Wallenberg Foundation. 

\bibliographystyle{splncs04}
\bibliography{reference}





\end{document}